\documentclass[11pt]{article}

\usepackage[final]{acl}

\usepackage{times}
\usepackage{latexsym}

\usepackage[T1]{fontenc}

\usepackage[utf8]{inputenc}

\usepackage{microtype}
\usepackage{booktabs} 
\usepackage{array} 
\usepackage{arydshln}
\usepackage{float} 
\usepackage{graphicx}
\usepackage{booktabs,subcaption}
\usepackage{caption}
\usepackage{subcaption}
\usepackage{xspace}

\usepackage{inconsolata}

\usepackage{multirow}

\def\ga{\texttt{ga}\xspace}
\def\gd{\texttt{gd}\xspace}
\def\qu{\texttt{qu}\xspace}

\def\NKdel#1{\bgroup\markoverwith{\textcolor{red}{\rule[0.4ex]{2pt}{3pt}}}\ULon{#1}}

\title{Modular Monolingual Adaptation using Pretrained Language Models}

\author{Nalin Kumar \and Ond\v rej Du\v sek  \\
  Charles University, Faculty of Mathematics and Physics \\
  Institute of Formal and Applied Linguistics\\
  Prague, Czechia \\
  \texttt{$\{$nkumar, odusek$\}$@ufal.mff.cuni.cz}}

\begin{document}
\maketitle
\begin{abstract}
Building monolingual language models (LMs) for low-resource languages typically relies on adapting pretrained language models (PLMs) by finetuning the whole model on the target language. This approach is widely favored over training from scratch, as it enables effective knowledge transfer.
Additionally, prior work has shown that using a language-specific tokenizer can enhance the adaptability. In this work, we hypothesize that full model tuning is often unnecessary and propose a more modular approach. Specifically, we replace the tokens, freeze the corresponding embeddings, and tune the rest of the model.
 We use Scottish Gaelic, Irish, and Quechua for our experiments, with Quechua being a very low-resource language (8.5k training instances). Evaluation on natural language understanding (NLU) tasks -- mask-filling, NER, and POS -- shows that our proposed approach improves performance when adapting the models to low-resource languages. Additionally, we provide a comprehensive analysis of the effectiveness of training strategies, the choice of pretrained embeddings, and models. 
 \end{abstract}

\section{Introduction}

There has been a great improvement in the performance of multilingual LMs in recent years, following the advent of large language models. They are not only resource-efficient and enable cross-lingual transfer learning for lower-resource languages, but also improve zero-shot skills. However, due to the \textit{curse of multilinguality} \cite{conneau-etal-2020-unsupervised}, when training a large multilingual model with fixed capacity, the higher resource languages might take up the major share of the parameter space, as models are predominantly trained on datasets consisting of English and other high-resource languages \cite{li2024quantifying,achiam2023gpt,team2023gemini,dubey2024llama,team2024gemma}. Consequently, this limits the downstream performance for underrepresented low-resource languages \cite{wu-dredze-2020-languages}. 

Previous literature suggests that, while training a model in several languages improves cross-lingual transfer, there might also be negative interference for low-resource languages \cite{wang-etal-2020-negative, muller-etal-2021-unseen}. There have been improvements with vocabulary extension \cite{chau-smith-2021-specializing}, adapters \cite{pfeiffer-etal-2020-mad}, and expert ensembles \cite{blevins-etal-2024-breaking}. One of the recent and most effective techniques involves language adaptive finetuning (LAFT)\cite{alabi2022adapting}, which incorporates training a pretrained multilingual LM (PMLM) on the same pretraining objective, such as masked language modeling \cite{eisenschlos-etal-2019-multifit}. However, training the whole model on the available little data might lead to overfitting and be too costly.

\begin{figure}
    \centering
    \includegraphics[width=1\linewidth]{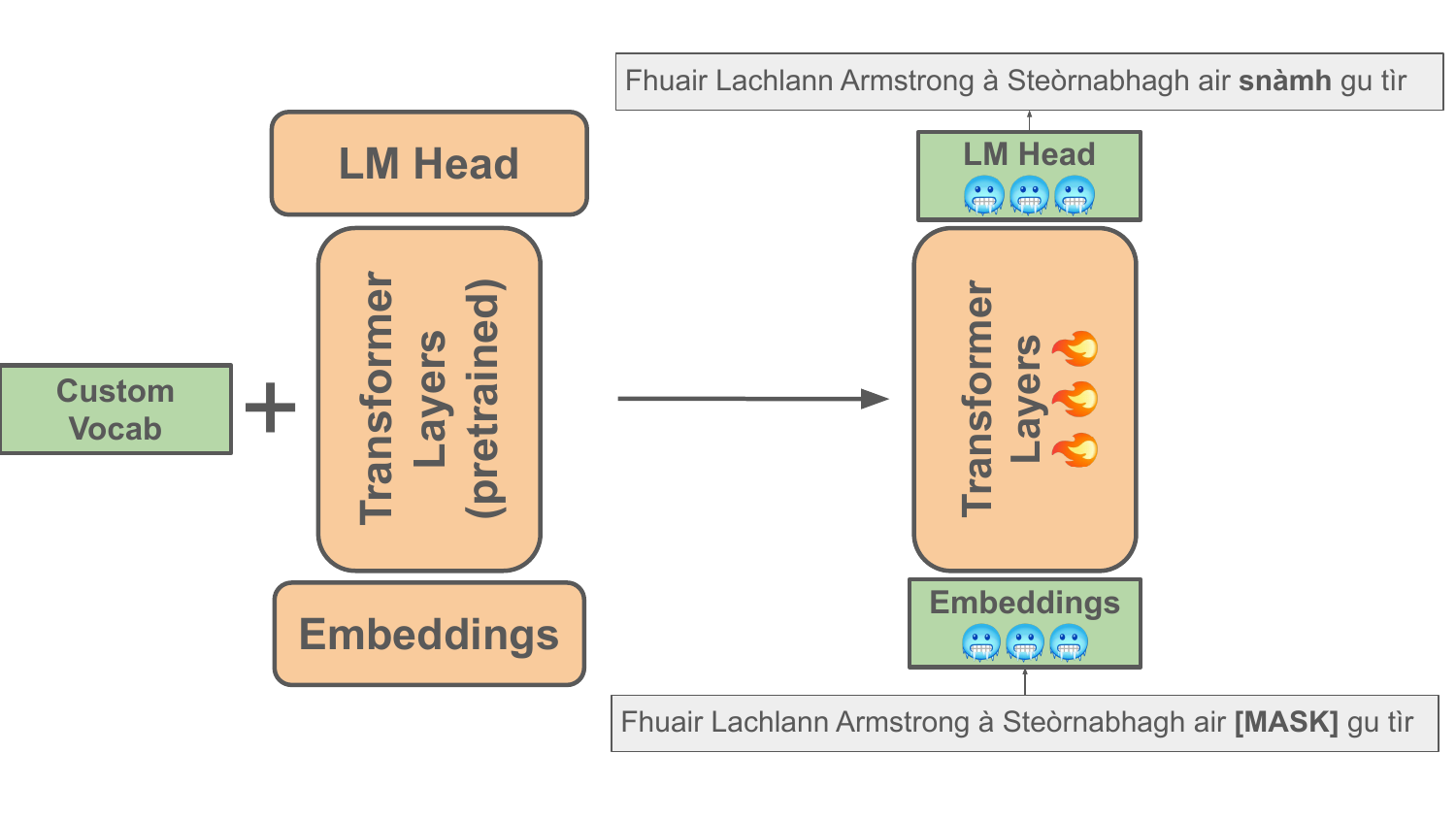}
 \caption{Overview of our proposed method. We first train a custom tokenizer and create corresponding embeddings using FastText. We replace the PLM's input and LM Head embeddings with the created embeddings and freeze them. We tune the rest of the model. As is standard in MLM models, the weights of input and output embeddings are tied.}
    \label{fig:approach}
\end{figure}

\begin{figure}[t]
    \centering
    \includegraphics[width=0.85\linewidth]{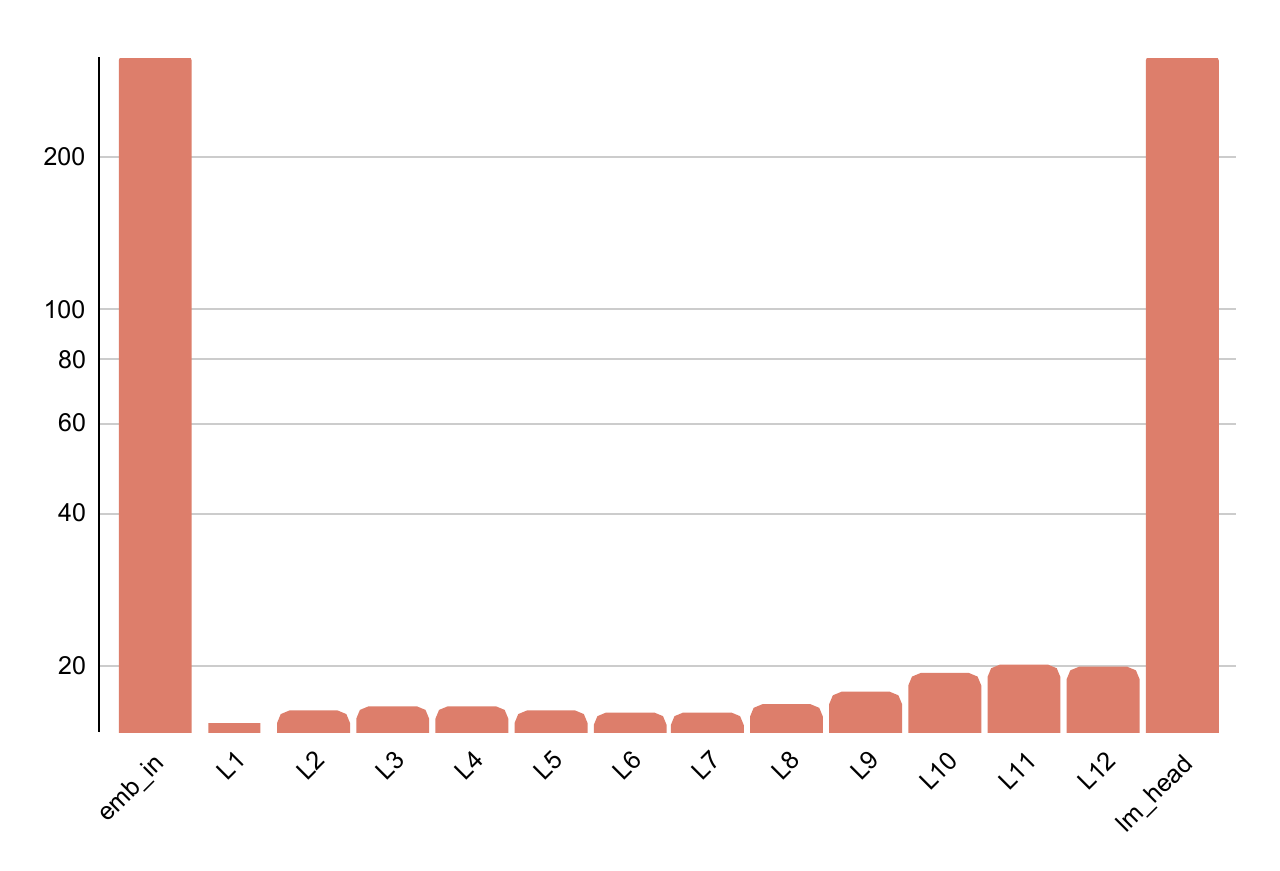}
    \caption{The graph shows the weight differences between each output layers of pretrained and finetuned BERT on Scottish Gaelic. The x-axis represents enumerated layers, while the y-axis represents the Euclidean distance between pretrained and finetuned BERT layers.}
    \label{fig:overfitting}
    \vspace{-1.5mm}
\end{figure}

In this work, we propose a modular adaptation of PMLMs using a custom tokenizer trained on a low-resource language. We freeze the embeddings, and train only the remaining parameters (see fig~\ref{fig:approach}). This prevents the model parameters from overfitting on the training examples. 
As is evident from Figure~\ref{fig:overfitting}, the embedding layers are particularly susceptible to overfitting due to a larger matrix size and larger weight updates during backpropagation. Thus, we hypothesize that training the whole model for monolingual adaptation is not always ideal in low-resource settings. 
Within the context of monolingual adaptation, we address the main research question (\textbf{RQ}) of our study: \textit{Does training the whole model give the best performance?} 

In particular, we make the following contributions:
\begin{itemize}
    \item We propose a modular framework for monolingual adaptation of pretrained multilingual language models (PMLMs) for low-resource languages. 
    \item We provide a comprehensive analysis on the role of tokenizers, choice of embedding initialization and effectiveness of multilingual models for monolingual adaptation. We perform experiments across three languages (Scottish Gaelic \gd, Irish \ga, and Quechua \qu) with multiple pretrained models (BERT, Multilingual BERT, Multilingual ModernBERT).
\end{itemize}


We evaluate our method on the mask-filling, named entity recognition (NER), and parts of speech (POS) tagging tasks. We show that 
the downstream performance of non-embedding tuning often surpasses the full tuning training strategy. However, the choice of embeddings plays a lesser role than anticipated. In contrast, having a custom tokenizer gives a major improvement over the multilingual tokenizer with a significantly larger vocabulary size. Our experimental code is released on GitHub.  \footnote{\url{https://github.com/knalin55/MMA-PLM}}


\section{Related Works}
\label{relatedWorks}

\paragraph{Role of vocabulary in Language Modeling}

The choice of vocabulary plays a vital role in training an LM for low-resource language. For instance, reusing the same vocabulary of a PMLM for an unseen language might result in increased token fertility, thereby increasing computational cost \cite{lundin2025token}. To mitigate this, significant work has been done to improve language modeling by scaling vocabulary size. \citet{chau-etal-2020-parsing} augment the vocabulary by replacing the [UNK] tokens in mBERT with 99 most frequent wordpiece tokens in the target language. They further train the model on the monolingual corpus of the target language to get better results than the baseline mBERT. Along a similar line of work, \citet{chau-smith-2021-specializing} proposed vocabulary augmentation and transliteration for languages with non-Latin scripts. In one of the recent works, \citet{yamaguchi2024can} experiment with vocabulary expansion in low-resource scenario. To augment the vocabulary more efficiently, \citet{lin-etal-2025-rethinking} use semantic and frequency-based method. However, \citet{limisiewicz-etal-2023-tokenization} suggest that vocabulary augmentation might not always help with all the downstream tasks, especially with token-level tasks. Another alternative is to exploit the current models and their embeddings. \citet{de-vries-nissim-2021-good} adapt English GPT2 model on Italian and Dutch by relearning only the lexical embeddings while freezing the transformer layers. \citet{hong-etal-2024-accelerating} use a language modeling head tailored to the target language and finetuning it further. In another work, \citet{rust-etal-2021-good} show that using a monolingual tokenizer works better than its multilingual counterpart. Building upon this work, we also experiment with replacing the pretrained model's vocabulary with the custom target language's tokens. 

\paragraph{LAFT}
Language Adaptive Finetuning (LAFT) method adapts pretrained language models to unseen languages by finetuning them on monolingual texts using the same pretraining objective, thereby improving performance and handling more linguistic nuances. \citet{sani-etal-2025-investigating} use LAFT for adapting AfriBERTa to Hausa, an extremely low-resource language. In another work, \citet{alabi-etal-2022-adapting} propose multilingual adaptive finetuning on several closely related languages. In recent years, Adapter-based finetuning \cite{pfeiffer-etal-2020-adapterhub} is one of the most cost-effective ways to train a model on a new language. 
However, despite having decent performance on target tasks, adapters have a worse generalizability over complete finetuning of models \cite{shuttleworth2024lora}.


\section{Proposed Approach} 
\label{approach}

We use encoder-based models such as BERT and mBERT \cite{devlin2019bert} for our experiments. To compare the effectiveness of models for monolingual adaptation on the basis of the presence of the target language in the pretraining data, we choose slightly older models. We use Masked Language Modeling (MLM) as our pretraining objective for training our encoder-based models. To address our \textbf{RQ}, we propose a more modular approach for monolingual adaptation (Figure~\ref{fig:approach}). We start with building a custom tokenizer for our target language using WordPiece \cite{devlin2019bert}, keeping the vocabulary size at 30k. We train the tokenizer on the same text corpus as the language model. This not only reduces the overall number of trainable parameters significantly (25$\%$ for mBERT), but also has better sub-word fertility for all models. To create the corresponding embedding matrix, we experiment with three different strategies:
\begin{itemize}
    \item Reusing base model embedding weights (model): We use the base tokenizer to re-tokenize all tokens in our newly created vocabulary. In general, most new language-specific tokens will be composed of multiple original model tokens. Therefore, we take the mean of the corresponding embedding weights. 
    \item Static pretrained embeddings (FastText): We use FastText \cite{bojanowski-etal-2017-enriching} to train the embeddings on the tokenized corpus (same corpus used for training tokenizer) with embedding dimension same as the model's. 
    \item Random: To compare the effectiveness of the choice of embeddings, we also experiment with randomly initialized embeddings. 
\end{itemize}
We initialize the model embeddings with the matrix created using one of the above strategies. To give better stability while training the models for low-resource languages, we freeze the input and output embeddings, and we then use the standard MLM objective to train only the non-embedding parameters on the target low-resource language. 


\section{Experimental Setup}
\label{expSetup}

\subsection{Datasets and Training Setup}
We run our experiments on the CC-100 dataset \cite{conneau-etal-2020-unsupervised,wenzek-etal-2020-ccnet} for the Scottish Gaelic (\gd), Irish (\ga), and Quechua (\qu) languages. 
The CC-100 dataset provides web-crawled unsupervised corpora for more than 100 languages. We preprocess the data by filtering out extremely small sentences. The resultant training data comprises 8.5k instances for \qu, 250k for \gd and 500k for \ga. We use this data to train our custom tokenizer, the FastText embeddings, and to train the language models with the MLM objective. 

We use the base variant for all our models. We train the models for 50 epochs with early stopping based on MLM accuracy. We use dynamic batch size depending on the available GPU. We use the default values for other training hyperparameters using HuggingFace Trainer. We report averaged scores over 3 pretraining runs, along with the standard deviation, for all settings.

\subsection{Evaluation Metrics}
We evaluate all our models on the mask-filling task (MLM).
We also use Named Entity Recognition (NER) and Parts-of-Speech Tagging (POS) as downstream tasks to evaluate our models. We use the WikiAnn dataset \cite{pan-etal-2017-cross,rahimi-etal-2019-massively,lovenia2024seacrowd} for the evaluation on the NER task. WikiAnn is a multilingual NER dataset consisting of annotations (LOC, PER, ORG) on Wikipedia articles. The \gd and \qu datasets comprise 100 instances for each training, development, and test set. The \ga dataset consists of 1000 instances each for train, development, and test sets.

Additionally, we use the UD Treebanks \cite{nivre-etal-2020-universal} as a source of data for the POS task. The \gd dataset consists of 3.5k, 656, and 548 instances for train, development, and test sets, respectively, with 17 unique tags from the universal POS tagset \cite{petrov-etal-2012-universal}. The train, development and test sets for \ga dataset consist of 4k, 451, and 454 instances. Due to the unavailability of a UD Treebank for the \qu language, we do not evaluate the POS tagging task.

We fully finetune the models on downstream tasks, as our goal here is to prioritize task-specific performance rather than generalization. Given that our experiments focus on extremely low-resource languages, we prioritize overall prediction correctness; therefore, we report scores using the accuracy metric across all 
considered tasks.


\begin{table}[t] 
\setlength{\tabcolsep}{4.5pt}        
\centering
\small
\begin{tabular}{llccc}
\toprule
 & \bf tokenization / & \multicolumn{3}{c}{\bf training} \\
 & \hfill \bf embeddings  &  full & emb & non-emb \\
\midrule
\multirow{5}{*}{\rotatebox{90}{\bf BERT}} & model / model & 21.8 $_{\pm 5.2}$ & \phantom{0}8.7 $_{\pm 3.5}$ & 23.0 $_{\pm 2.1}$ \\
 & model / random & 23.2 $_{\pm 0.7}$ & \phantom{0}0.0 $_{\pm 0.0}$ & 23.6 $_{\pm 0.0}$ \\
 & custom / model & 32.9 $_{\pm 1.4}$ & 21.8 $_{\pm 1.4}$ & 50.5 $_{\pm 0.4}$ \\
 & custom / FastText \hspace{-2mm} & 34.0 $_{\pm 5.7}$ & \phantom{0}1.5 $_{\pm 0.1}$ & 53.4 $_{\pm 0.2}$ \\
 & custom / random & 40.3 $_{\pm 0.6}$ & \phantom{0}0.0$_{\pm 0.0}$ & 49.3 $_{\pm 0.3}$ \\

\midrule
 & out-of-the-box model \hspace{-5mm} & \multicolumn{3}{c}{\phantom{0}6.9} \\
\multirow{5}{*}{\rotatebox{90}{\bf mBERT}} & model / model & 28.3 $_{\pm 2.2}$ & 24.1 $_{\pm 0.1}$ & 34.0 $_{\pm 1.0}$ \\
 & model / random & 29.7 $_{\pm 1.1}$ & \phantom{0}0.4 $_{\pm 0.6}$ & 21.0 $_{\pm 0.8}$ \\
 & custom / model & 33.4 $_{\pm 1.0}$ & 32.2 $_{\pm 0.1}$ & 54.3 $_{\pm 0.2}$ \\
 & custom / FastText \hspace{-2mm} & 42.4 $_{\pm 0.5}$ & \phantom{0}1.3 $_{\pm 0.1}$ & 55.0 $_{\pm 0.1}$ \\
 & custom / random & 38.6 $_{\pm 0.7}$ & \phantom{0}0.0 $_{\pm 0.1}$ & 49.5 $_{\pm 1.5}$ \\
\midrule

\multirow{5}{*}{\rotatebox{90}{\bf mmBERT}} & model / model & 30.9 $_{\pm 0.6}$ & 24.2 $_{\pm 0.4}$ & 41.3 $_{\pm 0.3}$ \\
 & model / random & 11.4 $_{\pm 1.5}$ & \phantom{0}0.1 $_{\pm 0.1}$ & 37.8 $_{\pm 0.0}$ \\
 & custom / model & 41.0 $_{\pm 1.5}$ & 19.8 $_{\pm 0.7}$ & \textbf{57.3 $_{\pm 0.6}$} \\
 & custom / FastText \hspace{-2mm} & \phantom{$_1$}20.6 $_{\pm 18.0}$ & \phantom{0}0.1 $_{\pm 0.0}$ & 54.1 $_{\pm 1.0}$\\
 & custom / random & 36.0 $_{\pm 1.4}$ & \phantom{0}0.1 $_{\pm 0.1}$ & 49.7 $_{\pm 0.4}$ \\

\midrule
\multirow{3}{*}{\rotatebox{90}{\bf rnd.~init.\hspace{-1mm}}} & model / random & 25.5 $_{\pm 0.8}$ & \phantom{0}0.0 $_{\pm 0.0}$ & 34.5 $_{\pm 1.0}$ \\
 & custom / FastText \hspace{-2mm} & 30.3 $_{\pm 0.2}$ & \phantom{0}0.0 $_{\pm 0.0}$ & 52.9 $_{\pm 0.4}$ \\
 & custom / random & 38.6 $_{\pm 0.1}$ & \phantom{0}0.7 $_{\pm 0.5}$ &  46.9 $_{\pm 0.2}$\\

\bottomrule
\end{tabular}
\caption{Mask filling accuracy for the Scottish Gaelic (\gd) language (rnd.~init.\ = randomly initialized model).}
\label{tab:gd:mfa}
\end{table}

\begin{table}[t]
\setlength{\tabcolsep}{4.5pt}
\centering
\small
\begin{tabular}{llccc}
\toprule
 & \bf tokenization / & \multicolumn{3}{c}{\bf training} \\
 & \hfill \bf embeddings  &  full & emb & non-emb \\
\midrule
\multirow{5}{*}{\rotatebox{90}{\bf BERT}} & model / model & 84.0 $_{\pm 0.6}$ & 80.6 $_{\pm 1.7}$ & \textbf{86.2 $_{\pm 1.2}$} \\
 & model / random & 83.1 $_{\pm 1.6}$ & 65.8 $_{\pm 3.7}$ & 83.1 $_{\pm 0.9}$ \\
 & custom / model & 84.7 $_{\pm 0.6}$ & 82.8 $_{\pm 0.9}$ & 85.0 $_{\pm 0.3}$ \\
 & custom / FastText \hspace{-2mm} & 82.4 $_{\pm 0.6}$ & 83.1 $_{\pm 0.8}$ & 75.5 $_{\pm 2.2}$ \\
 & custom / random & 81.6 $_{\pm 1.3}$ & 79.4 $_{\pm 0.6}$ & 80.2 $_{\pm 1.9}$ \\

\midrule

\multirow{5}{*}{\rotatebox{90}{\bf mBERT}} & model / model & 83.9 $_{\pm 1.1}$ & 83.1 $_{\pm 1.3}$ & 82.7 $_{\pm 0.4}$ \\
 & model / random & 81.2 $_{\pm 3.3}$ & 55.9 $_{\pm 0.6}$ & 81.1 $_{\pm 0.7}$ \\
 & custom / model & 85.3 $_{\pm 0.3}$ & 83.5 $_{\pm 1.5}$ & 85.0 $_{\pm 0.2}$ \\
 & custom / FastText \hspace{-2mm} & 84.3 $_{\pm 0.5}$ & 71.3 $_{\pm 0.4}$ & 84.2 $_{\pm 1.2}$ \\
 & custom / random & 82.1 $_{\pm 0.7}$ & 61.2 $_{\pm 5.8}$ & 80.7 $_{\pm 0.6}$ \\
\midrule

\multirow{5}{*}{\rotatebox{90}{\bf mmBERT}} & model / model & 63.4 $_{\pm 2.6}$ & 62.8 $_{\pm 2.4}$ & 63.1 $_{\pm 2.8}$ \\
 & model / random & 56.3 $_{\pm 3.2}$ & 54.8 $_{\pm 0.0}$ & 59.6 $_{\pm 1.3}$ \\
 & custom / model & 77.5 $_{\pm 1.7}$ & 81.8 $_{\pm 2.0}$ & 77.7 $_{\pm 0.9}$ \\
 & custom / FastText \hspace{-2mm} & 77.1 $_{\pm 2.5}$ & 75.6 $_{\pm 3.3}$ & 75.0 $_{\pm 1.9}$ \\
 & custom / random & 75.9 $_{\pm 0.9}$ & \phantom{$_1$}55.8 $_{\pm 12.2}$ & 76.6 $_{\pm 0.2}$ \\

\midrule

\multirow{3}{*}{\rotatebox{90}{\bf rnd.~init.\hspace{-1mm}}} & model / random & 82.3 $_{\pm 1.9}$ & 71.5 $_{\pm 0.6}$  & 80.4 $_{\pm 2.0}$ \\
 & custom / FastText \hspace{-2mm} & 79.8 $_{\pm 1.4}$ & 74.2 $_{\pm 1.0}$ & 80.0 $_{\pm 0.1}$\\
 & custom / random &  82.1 $_{\pm 0.4}$ &  76.1 $_{\pm 1.4}$ & 81.3 $_{\pm 1.0}$ \\
\bottomrule
\end{tabular}

\caption{NER accuracy in Scottish Gaelic (\gd).}
\label{tab:gd:ner}
\end{table}

\begin{table}[t]
\setlength{\tabcolsep}{5pt}
\centering
\small
\begin{tabular}{llccc}
\toprule
 & \bf tokenization / & \multicolumn{3}{c}{\bf training} \\
 & \hfill \bf embeddings  &  full & emb & non-emb \\
\midrule
\multirow{5}{*}{\rotatebox{90}{\bf BERT}} & model / model & 97.4 $_{\pm 0.1}$ & 95.5 $_{\pm 0.3}$ & 97.3 $_{\pm 0.0}$ \\
 & model / random & 96.6 $_{\pm 0.0}$ & 65.1 $_{\pm 0.2}$ & 96.3 $_{\pm 0.2}$ \\
 & custom / model & 96.9 $_{\pm 0.1}$ & 96.1 $_{\pm 0.1}$ & 96.9 $_{\pm 0.0}$ \\
 & custom / FastText \hspace{-2mm} & 96.7 $_{\pm 0.1}$ & 94.5 $_{\pm 0.2}$ & 96.7 $_{\pm 0.1}$ \\
 & custom / random & 96.1 $_{\pm 0.1}$ & 95.6 $_{\pm 0.2}$ & 96.0 $_{\pm 0.1}$ \\
\midrule

\multirow{5}{*}{\rotatebox{90}{\bf mBERT}} & model / model & 97.5 $_{\pm 0.1}$ & 97.0 $_{\pm 0.3}$ & \textbf{97.7 $_{\pm 0.1}$} \\
 & model / random & 96.2 $_{\pm 0.3}$ & 72.7 $_{\pm 0.0}$ & 96.3 $_{\pm 0.1}$ \\
 & custom / model & 97.2 $_{\pm 0.1}$ & 96.7 $_{\pm 0.0}$ & 97.2 $_{\pm 0.1}$ \\
 & custom / FastText \hspace{-2mm} & 97.0 $_{\pm 0.1}$ & 97.0 $_{\pm 0.0}$ & 93.9 $_{\pm 0.2}$ \\
 & custom / random & 96.4 $_{\pm 0.1}$ & 80.0 $_{\pm 0.3}$ & 96.3 $_{\pm 0.2}$ \\
\midrule

\multirow{5}{*}{\rotatebox{90}{\bf mmBERT}} & model / model & 89.6 $_{\pm 0.6}$ & 90.6 $_{\pm 0.4}$ & 90.6 $_{\pm 0.3}$ \\
 & model / random & 88.1 $_{\pm 0.4}$ & 89.2 $_{\pm 0.7}$ & 89.1 $_{\pm 0.6}$ \\
 & custom / model & 97.1 $_{\pm 0.1}$ & 97.0 $_{\pm 0.1}$ & 97.3 $_{\pm 0.1}$ \\
 & custom / FastText \hspace{-2mm} & 96.8 $_{\pm 0.1}$ & 94.2 $_{\pm 0.3}$ & 96.7 $_{\pm 0.1}$ \\
 & custom / random & 96.0 $_{\pm 0.1}$ & 88.3 $_{\pm 0.4}$ & 95.6 $_{\pm 0.1}$ \\

\midrule

\multirow{3}{*}{\rotatebox{90}{\bf rnd.~init.\hspace{-1mm}}} & model / random & 96.5 $_{\pm 0.2}$ & 72.9 $_{\pm 0.4}$ & 96.0 $_{\pm 0.2}$\\
 & custom / FastText \hspace{-2mm} & 95.8 $_{\pm 0.2}$ & 82.8 $_{\pm 0.3}$ &  96.0 $_{\pm 0.1}$\\
 & custom / random & 96.1 $_{\pm 0.1}$ & 81.5 $_{\pm 0.2}$ &  96.0 $_{\pm 0.2}$ \\

\bottomrule
\end{tabular}

\caption{POS tagging accuracy in Scottish Gaelic (\gd).}
\label{tab:gd:pos}
\end{table}

\subsection{Model Variants}

We use multilingual modern BERT (mmBERT) \cite{marone2025mmbert}, multilingual BERT (mBERT), and standard monolingual BERT \cite{devlin2019bert} as our base models. We also compare to a randomly initialized transformer model of the same shape as mBERT. 

We experiment with \emph{custom} tokenizers, along with the three choices of embeddings discussed in Section~\ref{approach} (\emph{model}/\emph{FastText}/\emph{random}). In addition, we compare these setups to re-using the models' own tokenization (\emph{model}), with either models' own or randomly initialized embeddings (\emph{model}/\emph{random}).

We apply the non-embedding training strategy (\emph{non-emb}), as described in Section~\ref{approach}. We compare it to training the full model (\emph{full}) or training the embedding parameters only (\emph{emb}) while keeping the internal transformer layers frozen.

Additionally, we compare our approach with the recent parameter-efficient finetuning method, low-rank adaptation (LoRA) \cite{hu2021loralowrankadaptationlarge}. The method decomposes each large matrix (i.e., all linear and attention layers of the model), into two small learnable low-rank matrices in the attention layer to reduce the training parameters. We use a LoRA rank of 64 and $\alpha$ = 128. 

\section{Results}
\label{res}
In this section, we provide a detailed analysis of our findings. In Tables~\ref{tab:gd:mfa}, \ref{tab:gd:ner} \& \ref{tab:gd:pos}, we report accuracy scores for mask-filling, NER and POS tagging tasks, respectively, on \gd using different models. We further compare our results with LoRA in Table~\ref{tab:gd:lora}. Table~\ref{tab:ga:mbert} reports the scores for the \ga language using mBERT, while Table~\ref{tab:qu:mbert} shows the results from mmBERT models trained for the \qu language. Additionally, we report basic system-level efficiency metric values in Table~\ref{tab:gd:sys}, comparing the efficiency of our best setup with full finetuning.

\paragraph{Training the whole model vs.~training non-em\-bedding parameters only}

The non-embedding training strategy consistently outperforms both full finetuning and embedding-only training on the mask-filling (MLM) task across all languages and experimental settings. This trend holds for \gd, \ga, and \qu, and is observed across different models and embedding initialization techniques. We attribute this improvement to the regularizing effect of freezing the input and output embedding layers. Updating the entire embedding matrix, which constitutes almost 25$\%$ (for masked language modeling) of the parameter space, can lead to overfitting, especially for low-resource languages. By freezing the embeddings and updating only the transformer layers, the model is able to adapt its higher-level representations without excessively altering the embedding space, resulting in better generalization.

For the NER and POS tagging downstream tasks, full finetuning and non-embedding training show comparable results, with both outperforming embedding-only training. This indicates that updating the embedding layer alone is insufficient for learning the task-specific contextual representations.

\paragraph{Effectiveness of using PLM}

Compared to a randomly initialized model in Table~\ref{tab:gd:mfa}, mBERT with FastText embeddings achieves a slightly higher MLM score (55.0) than the randomly initialized transformer (52.9). 

Similarly, for the NER and POS tagging tasks, mBERT-based models has better performance. These results demonstrate the benefits of pretrained language models and their ability to transfer knowledge effectively to low-resource settings.

\paragraph{Effectiveness of multilingual PLM}
We train a BERT model for similar settings and report the scores for the \gd language in Tables~\ref{tab:gd:mfa}, \ref{tab:gd:ner} \& \ref{tab:gd:pos} for MLM, NER and POS tagging tasks, respectively. Compared to mBERT and mmBERT, BERT achieves slightly lower overall MLM scores. In the default configuration (model/model), multilingual models consistently outperform BERT across all training strategies. However, when the embeddings are randomly initialized, the performance of BERT
slightly surpasses mBERT and mmBERT on the MLM task, showing the critical role of pretrained multilingual embeddings in enabling effective cross-lingual knowledge transfer.

For the NER task, BERT performs comparably to or slightly better than mBERT under embedding-only and non-embedding training setups. In contrast, with full finetuning, mBERT achieves higher NER scores than BERT. Meanwhile, mmBERT yields noticeably lower NER performance than BERT across all configurations.

\begin{table}[t]
\small
\setlength{\tabcolsep}{5pt}
\begin{tabular}{llccc}
\toprule
 & \bf tokenization / & \multicolumn{3}{c}{\bf training} \\
 & \hfill \bf embeddings  &  full & emb & non-emb \\
\midrule
 & out-of-the-box model\hspace{-8mm} & \multicolumn{3}{c}{21.2}\\
\multirow{5}{*}{\rotatebox{90}{\bf MLM}} & model / model & 28.3 $_{\pm 1.1}$ & 27.1 $_{\pm 0.2}$ & 39.4 $_{\pm 1.5}$ \\
 & model / random & 27.4 $_{\pm 0.8}$  & 00.3 $_{\pm 0.2}$ & 37.4 $_{\pm 1.3}$ \\
 & custom / model & 34.2 $_{\pm 3.4}$ & 38.0 $_{\pm 0.3}$ &  \textbf{57.1 $_{\pm 0.3}$}\\
 & custom / FastText \hspace{-2mm} & 13.1 $_{\pm 6.0}$ & 00.3 $_{\pm 0.1}$ &  46.8 $_{\pm 9.2}$\\
 & custom / random & 36.6 $_{\pm 2.3}$ & 00.6 $_{\pm 0.6}$ & 52.4 $_{\pm 0.8}$ \\

\midrule

\multirow{5}{*}{\rotatebox{90}{\bf NER}} & model / model & 93.1 $_{\pm 0.1}$ & 93.1 $_{\pm 0.1}$ & 93.0 $_{\pm 0.0}$\\
 & model / random & 91.0 $_{\pm 0.5}$ & 72.7 $_{\pm 1.6}$ & 91.1 $_{\pm 0.3}$\\
 & custom / model & 93.1 $_{\pm 0.6}$ & 92.1 $_{\pm 0.4}$ & \textbf{93.2 $_{\pm 0.3}$}\\
 & custom / FastText \hspace{-2mm} &  92.0 $_{\pm 0.4}$ & 88.3 $_{\pm 0.8}$ & 91.7 $_{\pm 0.4}$ \\
 & custom / random & 90.9 $_{\pm 0.2}$ & 71.3 $_{\pm 0.4}$  & 90.3 $_{\pm 0.0}$ \\
\midrule

\multirow{5}{*}{\rotatebox{90}{\bf POS}} & model / model & \textbf{96.7 $_{\pm 0.0}$} & 96.0 $_{\pm 0.0}$ & 96.1 $_{\pm 0.1}$ \\
 & model / random &  95.8 $_{\pm 0.2}$ & 75.2 $_{\pm 0.2}$ & 95.9 $_{\pm 0.3}$ \\
 & custom / model & 96.0 $_{\pm 0.3}$ & 95.2 $_{\pm 0.2}$ &  96.0 $_{\pm 0.1}$\\
 & custom / FastText \hspace{-2mm} & 95.8 $_{\pm 0.3}$ & 95.7 $_{\pm 0.2}$ & 93.3 $_{\pm 0.8}$\\
 & custom / random & 95.6 $_{\pm 0.1}$ & 83.0 $_{\pm 0.5}$ & 95.4 $_{\pm 0.2}$ \\
\bottomrule
\end{tabular}
\caption{Accuracy for mBERT on Irish (\ga) mask filling, POS and NER.}
\label{tab:ga:mbert}
\end{table}

\begin{table}[t]
\small
\setlength{\tabcolsep}{5pt}
\begin{tabular}{llccc}
\toprule
 & \bf tokenization / & \multicolumn{3}{c}{\bf training} \\
 & \hfill \bf embeddings  &  full & emb & non-emb \\
\midrule
\multirow{5}{*}{\rotatebox{90}{\bf MLM}} & model / model & 09.2 $_{\pm 0.3}$ & 02.3 $_{\pm 0.2}$ & 08.9 $_{\pm 0.1}$ \\
 & model / random & 06.0 $_{\pm 0.4}$ & 00.0 $_{\pm 0.0}$ & 07.8 $_{\pm 0.3}$ \\
 & custom / model & 24.0 $_{\pm 0.5}$ & 01.8 $_{\pm 0.1}$ & \textbf{29.4 $_{\pm 0.4}$} \\
 & custom / FastText \hspace{-2mm} & 23.7 $_{\pm 0.9}$ & 00.9 $_{\pm 0.2}$ & 28.5 $_{\pm 0.5}$ \\
 & custom / random & 18.7 $_{\pm 1.6}$ & 00.0 $_{\pm 0.0}$ &  23.0 $_{\pm 1.3}$\\

\midrule

\multirow{5}{*}{\rotatebox{90}{\bf NER}} & model / model & 63.2 $_{\pm 3.5}$ & 64.0 $_{\pm 2.7}$ & 59.6 $_{\pm 1.6}$\\
 & model / random & 56.0 $_{\pm 0.2}$ & 57.5 $_{\pm 0.6}$ & 58.0 $_{\pm 0.3}$\\
 & custom / model & \textbf{84.6 $_{\pm 0.7}$} & 81.9 $_{\pm 0.1}$ & 82.3 $_{\pm 1.4}$ \\
 & custom / FastText \hspace{-2mm} & 81.6 $_{\pm 2.6}$ & 78.3 $_{\pm 1.4}$ & 81.7 $_{\pm 0.9}$\\
 & custom / random & 64.9 $_{\pm 5.7}$ & 55.5 $_{\pm 3.2}$ &  69.7 $_{\pm 0.7}$\\

\bottomrule
\end{tabular}
\caption{Accuracy for mBERT on Quechua (\qu) mask filling and NER.}
\label{tab:qu:mbert}
\end{table}

\begin{table}[t]
\small
\setlength{\tabcolsep}{5pt}
\begin{tabular}{llccc}
\toprule
 & \bf tokenization / & \multicolumn{3}{c}{\bf task accuracy} \\
 & \hfill \bf embeddings  &  MLM & NER & POS \\
\midrule
 \vspace{0.5mm}\multirow{3}{*}{\rotatebox{90}{ \bf mBERT}} & model / model & 23.8 $_{\pm 1.2}$ & 79.1 $_{\pm 1.9}$ & 95.3 $_{\pm 0.1}$\\
 & custom / model & \textbf{35.1 $_{\pm 0.9}$} & \textbf{80.8 $_{\pm 0.9}$}  &  \textbf{95.8 $_{\pm 0.2}$} \\
 \vspace{0.5mm}& custom / FastText \hspace{-2mm} & 23.1 $_{\pm 1.3}$ & 70.1 $_{\pm 0.7}$  &  92.5 $_{\pm 0.3}$\\
\midrule

\multirow{3}{*}{\rotatebox{90}{\bf BERT}} & model / model & 16.3 $_{\pm 0.3}$ & 78.9 $_{\pm 0.0}$ & 94.5 $_{\pm 0.4}$ \\
 & custom / model & 31.9 $_{\pm 1.0}$ & 79.8 $_{\pm 1.3}$ &  95.1 $_{\pm 0.1}$\\
 & custom / FastText \hspace{-2mm} & 30.3 $_{\pm 0.9}$  & 74.0 $_{\pm 0.3}$ &  93.7 $_{\pm 0.1}$ \\

\bottomrule
\end{tabular}
\caption{Mask filling, NER and POS accuracy for LoRA finetuning on Scottish Gaelic (\gd).}
\label{tab:gd:lora}
\end{table}

\begin{table}[t]
\small
\setlength{\tabcolsep}{5pt}
\begin{tabular}{lcccc}
\toprule
 \hfill \bf model / method  &  \# tr. par. & VRAM & tr. time & latency \\
\midrule
 mBERT / full & 178M  & 1.3G  & 32H  &   9.7ms \\
 mBERT / ours & 85.6M  & 0.9G  & 14H  &   8.6ms\\

\midrule

 BERT / full & 108M  & 0.8G  & 30H  &   9.0ms \\
 BERT / ours & 85.6M  & 0.9G  & 15H  &   8.6ms \\

\midrule

 mmBERT / full & 308M  & 2.3G  & 71H  &   19.2ms \\
 mmBERT / ours & 110M  & 1.0G & 21H  &   13.3ms \\
\bottomrule
\end{tabular}
\caption{System-level efficiency metrics on Scottish Gallic (\gd), comparing full finetuning with our setup (custom tokenization, model-initialized embeddings, non-embedding finetuning). We report the number of trainable parameters (\# tr. par.) in million parameters (M), VRAM usage at training time in gigabytes (G), training time (tr. time) in hours (H), and average inference latency (time to process one input) in milliseconds (ms).}
\label{tab:gd:sys}
\end{table}

\paragraph{Language inclusion in pretraining data}
Table~\ref{tab:ga:mbert} provides MLM, NER and POS tagging task scores on mBERT for the \ga language. As shown in the table, full finetuning yields only a slight improvement in MLM performance over the out-of-the-box mBERT model, supporting the idea that a language's prior inclusion in the model's multilingual pretraining data facilitates monolingual adaptation to this language. 
The 
custom/model setting with non-embedding training still gives an improvement of more than 2.5$\times$ over the base model for the MLM task, regardless of the fact that the data set was already included in the pretraining setup.


\paragraph{Choice of tokenizer and embeddings}

In line with \citet{rust-etal-2021-good}, our custom tokenizer consistently outperforms the multilingual tokenizer on the mask-filling task. This improvement can be attributed to the improved finetuned subword segmentation on the target language, which reduces fertility and improves the proportion of continued words. In addition to improving performance, the custom tokenizer also substantially reduces model size -- particularly for token classification tasks -- since the embedding layers occupy nearly 25$\%$ of the total parameters. A smaller, language-specific vocabulary, thereby, not only enhances efficiency but also increases training speed due to lower computational and memory requirements.

In contrast, the choice of embedding initialization does not play a major role in the non-embedding training setting. There is only a slight improvement in the MLM scores for FastText embeddings over the other choices (Table~\ref{tab:gd:mfa}). Randomly initialized embeddings, interestingly, perform only slightly worse than both model-based and FastText embeddings. This suggests that the transformer layers can relearn and realign to the new embedding space, acting as an effective regularization strategy. Under full finetuning, however, the default embeddings seem to have a negative impact, likely because the model is heavily biased toward non-\gd knowledge and has to first unlearn irrelevant representations. 

\paragraph{Better encoder model}
mmBERT achieves slightly better performance than mBERT and BERT on the mask-filling task. This gain can likely be attributed to its larger model capacity and stronger cross-lingual representations, due to training on a more diverse pretraining corpus. Due to its richer embedding space, mmBERT also outperforms the corresponding FastText-based setting. 
Surprisingly, the performance of mmBERT is lower than mBERT and BERT in some variants for the downstream tasks. We suspect that the reason might be “catastrophic overtraining” \cite{springer2025overtrained}, due to which the model fails to finetune on a downstream task. 

\paragraph{Training on very low-resource language}
Experiments for \qu follow a similar trend as for \gd. In particular, the non-embedding training strategy outperforms full tuning with default and FastText emebddings. Furthermore, similar to previous trends, models trained with the default tokenizer underperform compared to those using a custom tokenizer. This can be attributed to the minimal representation of \qu in the model’s pretraining data, resulting in weaker language representations.

In the NER evaluation, models with default tokenization again performs worse than their custom-tokenized counterparts. Among the custom tokenizer variants, FastText and model embeddings have comparable performance, while random embeddings perform slightly worse than both.

Overall, the performance on \qu remains low across both evaluation metrics, primarily due to the limited size of the available training data. 

\paragraph{Comparison with LoRA}
Although LoRA scores are comparable to the full finetuning strategy, our proposed training approach consistently outperforms LoRA on all three tasks (Table~\ref{tab:gd:lora}). We suspect that the method's low rank restricts the model's capacity to learn more complex tasks, thereby limiting performance. 

\paragraph{System-level metrics}
As shown in Table~\ref{tab:gd:sys}, our proposed approach is more energy-efficient than model finetuning. Since our method has a smaller vocabulary, the number of training parameters is significantly lower than in full model tuning, thereby using less VRAM and having better latency at inference time. Additionally, due to better embedding initialization and smaller model size, our approach converges faster than the conventional full training method.  

\section{Conclusion}
\label{conc}

In this work, we adapt a PLM to a new language. We hypothesize that tuning the entire parameter space is often unnecessary and might even be suboptimal for low-resource language modeling. To address this, we propose a modular monolingual adaptation strategy that replaces the tokenizer with a language-specific one, initializes new embeddings using different strategies, freezes the embedding layers, and trains only the non-embedding parameters using the masked language modeling objective. 

We experiment with multiple models (BERT, mBERT, and mmBERT) in multiple languages (\gd, \ga $\&$ \qu) and training strategies. Our results consistently show that non-embedding training outperforms or matches full finetuning on the masked language modeling task, while requiring fewer trainable parameters and reducing training cost. We attribute this improvement to better regularization due to frozen embeddings, which mitigates overfitting in low-resource settings. On downstream NER and POS tagging tasks, non-embedding training performs comparably to full finetuning and consistently outperforms embedding-only training, thereby showing the effectiveness of our proposed approach. 

We show that multilingual pretrained models provide clear advantages over randomly initialized transformers, demonstrating effective cross-lingual knowledge transfer—even when the target language is absent or minimally represented in pretraining. Furthermore, the inclusion of a related language in pretraining results in effective adaptation, but it does not seem to be a necessary condition. Also, following previous work, we validate the use of a custom tokenizer over multilingual tokenizers. It consistently improves performance over the default multilingual tokenizer and reduces fertility and improves the proportion of continued words, particularly for low-resource languages. In contrast, the choice of embedding initialization (model-based, FastText, or random) has a relatively minor performance improvement for the non-embedding training settings, suggesting that the transformer layers can effectively realign to new lexical spaces. 

Overall, our findings demonstrate that full model finetuning is not always the optimal strategy for low-resource monolingual adaptation. A modular approach that combines custom tokenization with selective parameter training provides a simple, parameter-efficient, and effective alternative. For future work, our approach can be extended to decoder-based architectures, more low-resource languages and building mutlilingual models for a language family.

\section{Limitations}
Our experiments cover only three low-resource languages (Scottish Gaelic, Irish, and Quechua), which limits the extent to which our approach transfers to typologically diverse languages with different scripts or morphological properties. The evaluation is restricted to relatively simple NLU tasks (mask-filling, NER, POS) due to the lack of available gold-standard data for more complex tasks in the target languages. While it is possible to generate synthetic data for such tasks based on English benchmarks, the result will induce noise due to alignment errors and localization/cultural biases. We reserve these experiments for further work.

\section{Acknowledgments}
This work was funded by the European Union (ERC, NG-NLG, 101039303) and by Charles University projects GAUK~302425 and SVV~260~821. It used resources of the LINDAT/\hspace{0mm}CLARIAH-CZ Research Infrastructure (Czech Ministry of Education, Youth, and Sports project No. LM2023062).
\bibliography{custom}

\appendix

\end{document}